\documentclass[letterpaper]{article}
\usepackage{aaai19}
\usepackage{times}
\usepackage{helvet}
\usepackage{courier}
\usepackage{url}
\usepackage{graphicx}
\usepackage{amsmath}
\usepackage{amssymb}
\usepackage{pbox}
\usepackage{booktabs}
\usepackage{microtype}
\usepackage{subcaption}
\title{DeepLogic: Towards End-to-End Differentiable Logical Reasoning}
\author{
Nuri Cingillioglu \and Alessandra Russo\\
\{nuric, a.russo\}@imperial.ac.uk\\
Deparment of Computing\\
Imperial College London
}
\begin{document}
\maketitle
\begin{abstract}
Combining machine learning with logic-based expert systems in order to get the best of both worlds are becoming increasingly popular. However, to what extent machine learning can already learn to reason over rule-based knowledge is still an open problem. In this paper, we explore how symbolic logic, defined as logic programs at a character level, is learned to be represented in a high-dimensional vector space using RNN-based iterative neural networks to perform reasoning. We create a new dataset that defines 12 classes of logic programs exemplifying increased level of complexity of logical reasoning and train the networks in an end-to-end fashion to learn whether a logic program entails a given query. We analyse how learning the inference algorithm gives rise to representations of atoms, literals and rules within logic programs and evaluate against increasing lengths of predicate and constant symbols as well as increasing steps of multi-hop reasoning.
\end{abstract}

\section{Introduction}
There has been increasing interest and attempts at combining machine learning with logic-based expert systems in order to get \emph{the best of both worlds}: have an end-to-end trainable system that requires little data to learn how to reason, possibly using existing background knowledge, and is interpretable by humans~\cite{neurosymsurvey}. However, it is not clear to what extent machine learning can learn logic-based reasoning without requiring any prior engineering. Thus, in this work we provide a first insight into how symbolic logic, represented as logic programs at a character level can be learned using RNN-based iterative neural networks to perform reasoning.

A crucial component in existing logic-based expert systems is the ability to iteratively use a given knowledge base. Framed as multi-hop reasoning, these tasks require an agent to process information over several steps to reach a conclusion. This paradigm of reasoning has been performed by neural networks in domains such as story comprehension with Dynamic Memory Networks~\cite{dmnvisual}, graph problems with Differentiable Neural Computers~\cite{dnc} and visual question answering with Relation Networks~\cite{relationnetworks} and Memory Attention Control Networks~\cite{mac}. Traditionally, inference systems over logic programs are manually built on algorithms such as backward chaining~\cite{bacwardchaining} or forward chaining~\cite{russell2016artificial}. There have been attempts to partially replace symbolic components of these systems with neural networks such as Neural Prolog~\cite{neuralprolog} which constructs networks through symbolic applications of logic rules. Another vital component, unification of variables have been tackled by Unification Neural Networks~\cite{neuralunification}. However, neither of these networks act as a complete inference engine that can be trained end-to-end solely using examples and learn corresponding representations of symbolic knowledge.
%Multi-hop reasoning is often expressed as applications of rules in logic~\cite{russell2016artificial} whereby the inference mechanism allows reasoning of the form ``human(X)$\rightarrow$mortal(X) $\land$ human(socrates) $\vdash$ mortal(socrates)".% and seeks to derive conclusions from a set of rules. %Can we thus replace the inference engine of a normal logic program with a neural network to learn not only representations of logic programs but also how to perform logical reasoning?

Another important aspect of machine learning that allows scalability is learning dense distributional representations such as word embeddings~\cite{glove}. This approach is used to learn embeddings of predicates to perform knowledge base completion~\cite{logictensor}; however, an existing reasoning mechanism such as backward chaining~\cite{timntp} guides the machine learning component to learn such embeddings. We instead build on the prior work on iterative neural networks that learn algorithmic tasks such as addition of digits~\cite{neuralturing} and sequence processing~\cite{learnsimplealgo}. Resultantly, we learn reasoning tasks from scratch without prior engineering alongside representations of knowledge bases expressed as logic programs. The key motivation is that if we can learn symbolic reasoning using machine learning then we might be able to minimise the extent to which they have to be combined manually. The focus is not to extract the prior symbolic hypotheses but to learn how to utilise a given knowledge base; thus, we do not compare within the domain of inductive logic programming~\cite{ilp}.

This paper provides the following contributions, (i) a new synthetic dataset consisting of 12 various classes of normal logic programs, (ii) an iterative neural inference network to learn end-to-end different reasoning tasks and (iii) analysis of how these networks represent logic programs and handle multi-hop reasoning. Our implementation is publicly available online at \url{https://github.com/nuric/deeplogic}.

\begin{table*}[ht]
\small
\centering
\caption{Sample programs from tasks 1 to 7.}
\label{tab:tasks17}
\begin{tabular}{@{}lllllll@{}}
\toprule
1: Facts & 2: Unification & 3: 1 Step & 4: 2 Steps & 5: 3 Steps & 6: Logical AND & 7: Logical OR\\
\midrule
\pbox{2cm}{e(l).\\
i(u).\\
n(d).\\
v(h,y).\\
p(n).}
& \pbox{2cm}{o(V,V).\\
i(x,z).\\
y(w,d).\\
p(a,b).\\
t(A,U).}
& \pbox{3cm}{g(L,S) :- x(S,L).\\
x(a,m).\\
y(X) :- r(X).\\
p(h).\\
s(t,v).}
& \pbox{3cm}{x(G,B) :- k(G,B).\\
k(Z,V) :- g(Z,V).\\
g(k,k).\\
e(k,s).\\
p(L,G) :- v(G,L).}
& \pbox{3cm}{p(P,R) :- b(R,P).\\
b(A,L) :- a(A,L).\\
a(W,F) :- v(F,W).\\
v(t,i).\\
l(D) :- t(D).}
& \pbox{3cm}{f(P,U) :- b(P) , p(U).\\
b(x).\\
p(x).\\
p(y).\\
e(y,v).}
& \pbox{3cm}{e(D,X) :- n(D,X).\\
e(D,X) :- w(D,X).\\
e(w,y).\\
n(n,j).\\
w(t).}\\
\midrule
\pbox{2cm}{? e(l). 1\\ ? i(d). 0}
& \pbox{2cm}{? o(d,d). 1\\ ? o(b,d). 0}
& \pbox{3cm}{? g(m,a). 1\\ ? g(a,m). 0}
& \pbox{3cm}{? x(k,k). 1\\ ? x(k,s). 0}
& \pbox{3cm}{? p(t,i). 1\\ ? p(i,t). 0}
& \pbox{3cm}{? f(x,x). 1\\ ? f(y,x). 0}
& \pbox{3cm}{? e(n,j). 1\\ ? e(i,j). 0}\\
\bottomrule
\end{tabular}
\end{table*}

\section{Background}\label{sec:background}
A \emph{normal logic program}~\cite{logicprogramming} is a set of rules of the form:
\begin{equation}\label{eq:clause}
A \leftarrow L_1, \ldots, L_n\ (n\geq0)
\end{equation}
where $A$ is an \emph{atom} and each $L_i$ is a literal. A \emph{literal} is either an atom (a positive literal) or a negated atom of the form $\texttt{not}\ A$ where \texttt{not} is \emph{negation by failure}~\cite{nbf}. Atom $A$ represents the \emph{head} and $L_1,\ldots,L_n$ the \emph{body} of a rule. An atom is a predicate with some arity, e.g. $p(X,Y)$, $q(a)$, where variables are upper case characters and predicates and constants are lower case. If a rule does not contain any variables we refer to it as \emph{ground} rule and ground rules with empty bodies as \emph{facts}. We follow a similar syntax to Prolog~\cite{prolog} and express normal logic programs as:
\begin{align}
p(X)\ &\text{:-}\ q(X),\ -r(X). \label{eq:rule} \\
q(a).& \label{eq:fact}
\end{align}
As in equation~\ref{eq:rule}, the $\leftarrow$ is replaced with $\text{:-}$ and the negation by failure \texttt{not} with $-$ while maintaining the same semantics. When there are no body literals we omit the implication entirely, equation~\ref{eq:fact}.
\begin{align}
%f : \mathcal{C} \times \mathcal{Q} &\rightarrow \mathbb{B} \label{eq:objective}\\
f(C,Q) &= \begin{cases}
            1 & \text{if}\ C \vdash Q \\
            0 & \text{otherwise}
          \end{cases} \label{eq:objective}\\
f(C,\operatorname{not} Q) &= 1 - f(C,Q) \label{eq:neg}
\end{align}
We define the logical reasoning process as an inference function $f$, equation~\ref{eq:objective}, that given a normal logic program (without function symbols) as context $C \in \mathcal{C}$ and a ground atom as query $Q \in \mathcal{Q}$ returns $\{1, 0\} = \mathbb{B}$ depending on whether the context entails the query $C \vdash Q$ or not. We can now define negation by failure $\operatorname{not} Q$ using the outcome of the corresponding positive ground query $Q$, equation~\ref{eq:neg}.
\begin{equation}\label{eq:nobjective}
f(C,Q)\ \triangleq\ p(Q|C)
\end{equation}
In order to learn the reasoning process, we define an auxiliary objective for the neural network and consider the inference function $f$ to be the conditional probability of the query given the context, equation~\ref{eq:nobjective}. This approach renders the problem, from a machine learning perspective, as a binary classification problem and allows training using standard cross-entropy loss.

\begin{table*}[ht]
\small
\centering
\caption{Sample programs from tasks 8 to 12.}
\label{tab:tasks812}
\begin{tabular}{@{}lllll@{}}
\toprule
8: Transitivity & 9: 1 Step NBF & 10: 2 Step NBF & 11: AND NBF & 12: OR NBF\\
\midrule
\pbox{4cm}{f(A,W) :- q(A,P) , d(P,W).\\
q(h,t).\\
d(t,j).\\
q(d,m).\\
d(n,g).\\
s(S,F) :- x(S,A) , e(A,F).}
& \pbox{3cm}{s(X,J) :- -p(J,X).\\
p(e,x).\\
v(V,Q) :- u(V,Q).\\
o(N) :- -q(N).\\
t(x,e).\\
m(y,c).}
& \pbox{3cm}{r(C) :-\ -o(C).\\
o(P) :- l(P).\\
l(o).\\
g(u).\\
p(U,L) :- e(U,L).\\
p(X,X).}
& \pbox{4cm}{b(G,B) :- -i(G) , u(B).\\
i(w).\\
g(a).\\
u(a).\\
f(t).\\
l(W) :- a(W) , d(W).}
& \pbox{4cm}{y(Z) :- -e(Z).\\
y(Z) :- b(Z).\\
y(r).\\
e(d).\\
s(a).\\
b(m).}\\
\midrule
\pbox{4cm}{? f(h,j). 1\\ ? f(d,g). 0}
& \pbox{3cm}{? s(x,e). 0\\ ? s(e,x). 1}
& \pbox{3cm}{? r(u). 1\\ ? r(o). 0}
& \pbox{3cm}{? b(a,a). 1\\ ? b(w,a). 0}
& \pbox{3cm}{? y(a). 1\\ ? y(d). 0}\\
\bottomrule
\end{tabular}
\end{table*}

\section{The Tasks}\label{sec:tasks}
Inspired by the bAbI dataset~\cite{babi}, we layout 12 tasks that cover various aspects of normal logic programs. The tasks are of increasing complexity building up on concepts seen in prior tasks. Every task consists of triples $(context, query, target)$ following the signature of the inference function in equation~\ref{eq:objective}. For compactness a single context can have multiple queries and are expressed in the form \texttt{?~query~target} after the context. The contexts contain the simplest possible rules that cover the required reasoning procedure and are not mixed together akin to unit tests in software engineering.

Each task is generated using a fixed procedure from which samples are drawn. For constants and predicates we use the lower case English alphabet and for variables upper case. The length of the character sequences that make up predicates, constants and variables can be of arbitrary length (only examples of length 1 are shown) but we generate lengths up to 2 for the training dataset and longer lengths for test data. The arity of the atoms are selected randomly between 1 and 2. For every sample we also generate irrelevant rules as \emph{noise} that always have different predicates and random structure while still preserving the semantics of the task.

\textbf{Facts} The simplest task consists only of facts. There is only one successful case in which the query appears in the context and can fail in 3 different ways: (i) the constant might not match (shown in Table~\ref{tab:tasks17}), (ii) the predicate might not match or (iii) the query might not be in the context at all. These failures can cascade and a query can fail for multiple reasons with equal probability.

\textbf{Unification} These tasks contain rules with empty bodies and an atom with variables in the head. The intended objective is to emphasise the semantics of unification between different $p(X,Y)$ and same variables $p(X,X)$. The query succeeds if the corresponding variables unify and fail otherwise. The failure case in which same variables do not match different constants is in Table~\ref{tab:tasks17}.

\textbf{N Step Deduction} One prominent feature of logical reasoning is the application of rules that contain literals in the body. These tasks cover the case when there is a single positive atom in the body. All such rules contain only variables and chains of arbitrary steps can be generated. For the training dataset, we generate up to 3 steps, samples in Table~\ref{tab:tasks17}. The query succeeds when the body of the last rule in the chain is grounded with the same constant as in the query. We occasionally swap the variables $p(X,Y)\text{:-}q(Y,X).$ to emphasise the variable binding aspect of rules which can happen at any rule in the chain or not at all. The failure cases are covered when the swapped constants do not match or when the final body literal in the chain fails due to reasons covered in the first task.

\textbf{Logical AND} Building upon the previous deduction tasks, we create rules with 2 body literals to capture the logical $\land$ semantics in rules, sample in Table~\ref{tab:tasks17}. The reasoning engine now has to keep track of and prove both body literals to succeed. A failure occurs when one randomly selected body literal fails for reasons similar to the first task.

\textbf{Logical OR} Having multiple matching heads captures the semantics of $\lor$ in logic programming by creating several possible paths to a successful proof, sample in Table~\ref{tab:tasks17}. In this task we branch the query predicate 3 ways, 2 implications and 1 ground rule. The query succeeds when any of the rules succeed and fail when all the matching rules fail.

\textbf{Transitivity} Based on the previous 2 tasks, the transitive case covers existential variable binding. It requires the model to represent the conjunction of the body literals of a rule and match multiple possible facts. The case succeeds when the inner variable unifies with the ground instances or fails otherwise. We expect an reasoning engine to solve the previous 2 tasks in order to solve this one, sample in Table~\ref{tab:tasks812}.

\textbf{N-Step Deduction with Negation} These tasks introduce the concept of \emph{negation by failure}. The body literal of the first rule in the chain is negated and a chain of arbitrary length is generated. For the training dataset we only generate proof chains of length 2, samples in Table~\ref{tab:tasks812}. The query succeeds when the negated body atom fails; the swapped variables do not match the constants, or the body literal of the final rule in the chain fails for reasons similar to the first task. The query fails whenever the negated body atom succeeds following the semantics described by equation~\ref{eq:neg}.

\textbf{Logical AND with Negation} After introducing negation by failure, we consider negation together with logical $\land$ and randomly negate one of the body literals, sample in Table~\ref{tab:tasks812}. As such, the query succeeds when the negated body atom fails and the other literal succeeds. The query can fail when either body literal fails similar to the non-negated case.

\textbf{Logical OR with Negation} Finally, we consider negation together with the logical $\lor$ case and negate the body literal of one rule. The query succeeds when any matching rule except the negated one succeeds and fails if the negated rule succeeds while other matching rules fail, sample in Table~\ref{tab:tasks812}.

\section{Neural Reasoning Networks}\label{sec:model}
In this section we describe a RNN-based iterative neural network for learning the inference function $f$, equation~\ref{eq:objective}. Broadly, we call networks that learn logical reasoning, Neural Reasoning Networks and the objective is to learn how $C \vdash Q$ is computed solely from examples. The primary challenge for these networks is to have a \emph{fixed} network architecture that must process all the tasks unlike tree based recurrent networks~\cite{treelstm} or graph networks~\cite{graphnetworks} which construct a different network dependant on each input. We place this constraint to avoid engineering any prior structural information into a network. We gather inspiration from Memory Networks~\cite{memnn}, in particular we wanted to incorporate the end-to-end approach~\cite{memn2n} and the iterative fashion of Dynamic Memory Networks (DMN)~\cite{dmn} while following the steps of a symbolic reasoning method such as backward~chaining~\cite{bacwardchaining}.
\begin{figure}[h]
\centering
\includegraphics[width=0.8\columnwidth]{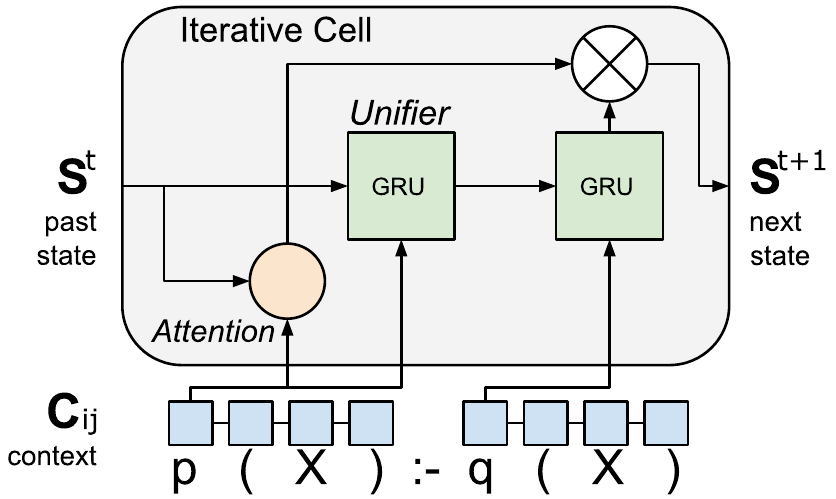}
\caption{Graphical overview of the iterative cell of the Iterative Memory Attention (IMA) model. The context and query are processed at the character level to produce literal embeddings, then an attention is computed over the head of the rules. A weighted sum of the unifier GRU outputs using the attention, updates the state for the next iteration.}
\label{fig:ima_diag}
\end{figure}

\begin{table*}[ht]
\small
\centering
\caption{Results on easy test set (10k each), $d=64$, $sm=\operatorname{softmax}$.}
\label{tab:results}
\begin{tabular}{@{}rccccccccccc@{}}
\toprule
Training  & \multicolumn{6}{c}{Multi-task} & \multicolumn{5}{c}{Curriculum} \\
Model     & LSTM & MAC & DMN & \multicolumn{3}{c}{IMA}  & MAC & DMN  & \multicolumn{3}{c}{IMA} \\
Embedding & - & rule & rule & \multicolumn{2}{c}{literal} & lit+rule & rule & rule & \multicolumn{2}{c}{literal} & lit+rule \\
Attention & - & $sm$ & $\sigma$ & $\sigma$ & $sm$ & $sm$ & $sm$ & $\sigma$ & $\sigma$ & $sm$ & $sm$ \\
\midrule
Facts       & 0.61 & 0.84 & 1.00 & 1.00 & 1.00 & 0.98 & 0.89 & 1.00 & 1.00 & 0.99 & 0.94 \\
Unification & 0.53 & 0.86 & 0.87 & 0.90 & 0.87 & 0.85 & 0.83 & 0.85 & 0.88 & 0.88 & 0.86 \\
1 Step      & 0.57 & 0.90 & 0.74 & 0.98 & 0.94 & 0.95 & 0.77 & 0.62 & 0.96 & 0.93 & 0.92 \\
2 Steps     & 0.56 & 0.81 & 0.67 & 0.95 & 0.95 & 0.94 & 0.70 & 0.58 & 0.95 & 0.91 & 0.89 \\
3 Steps     & 0.57 & 0.78 & 0.77 & 0.94 & 0.94 & 0.94 & 0.64 & 0.64 & 0.93 & 0.86 & 0.87 \\
AND         & 0.65 & 0.84 & 0.80 & 0.95 & 0.94 & 0.85 & 0.81 & 0.70 & 0.80 & 0.78 & 0.83 \\
OR          & 0.62 & 0.85 & 0.87 & 0.97 & 0.96 & 0.93 & 0.75 & 0.75 & 0.96 & 0.93 & 0.90 \\
Transitivity& 0.50 & 0.50 & 0.50 & 0.50 & 0.52 & 0.52 & 0.50 & 0.50 & 0.50 & 0.50 & 0.50 \\
1 Step NBF  & 0.58 & 0.92 & 0.79 & 0.98 & 0.94 & 0.95 & 0.65 & 0.58 & 0.96 & 0.91 & 0.92 \\
2 Steps NBF & 0.57 & 0.83 & 0.85 & 0.96 & 0.93 & 0.95 & 0.57 & 0.73 & 0.95 & 0.90 & 0.90 \\
AND NBF     & 0.55 & 0.82 & 0.84 & 0.92 & 0.93 & 0.85 & 0.61 & 0.61 & 0.71 & 0.77 & 0.83 \\
OR NBF      & 0.53 & 0.74 & 0.75 & 0.86 & 0.86 & 0.86 & 0.59 & 0.63 & 0.86 & 0.83 & 0.84 \\
\midrule              
Easy Mean   & 0.57 & 0.81 & 0.79 & \textbf{0.91} & 0.90 & 0.88 & 0.69 & 0.68 & 0.87 & 0.85 & 0.85 \\
Medium Mean & 0.52 & 0.70 & 0.70 & \textbf{0.86} & 0.81 & 0.79 & 0.60 & 0.61 & 0.81 & 0.76 & 0.74 \\
Hard Mean   & 0.51 & 0.63 & 0.66 & \textbf{0.83} & 0.75 & 0.72 & 0.55 & 0.58 & 0.76 & 0.70 & 0.68 \\
\bottomrule
\end{tabular}
\end{table*}
\textbf{Design} We can consider the logic program context a read-only memory and the proof state a writable memory component. In a similar fashion to Prolog's backward chaining algorithm~\cite{prolog}, we aim to have (i) a state to store information about the proof such as the query, (ii) a mechanism to select rules via attention and (iii) a component to update the state with respect to the rules. To that end, we introduce the Iterative Memory Attention (IMA) network that given a normal logic program as context and a positive ground atom as query, embeds the literals in a high dimensional vector space, attends to rules using soft attention and updates the state using a recurrent network. IMA should be considered a variant of Memory Networks, designed to suit the logical reasoning process in which the recurrent component is iterated over literals to perform reasoning, a graphical overview is shown in Figure~\ref{fig:ima_diag}. 

\subsection{Literal Embedding}
The inputs to the network are two sequences of characters $c^C_0, \ldots, c^C_m$ and $c^Q_0, \dots, c^Q_n$ for context and query respectively. The principle motivation behind having character level inputs rather than symbol tokens is to constrain the network to learn sub-symbolic representations that could potentially extend to previously unseen literals. We pre-process the context sequence to separate it into literals and obtain an input tensor $\mathbf{I}^C \in \mathbb{N}^{R \times L \times m'}$ of characters encoded as positive integers where $R$ is the number of rules, $L$ number of literals and $m'$ the length of the literals. The query is a single ground positive atom encoded as a vector $I^Q \in \mathbb{N}^n$. This pre-processing allows the network to consider each literal independently when iterating over rules giving it finer control over the reasoning process. 
\begin{equation}\label{eq:lit_embed}
h_t = GRU(O[\mathbf{I}_{::t}], h_{t-1})
\end{equation}
Each literal is embedded using a recurrent neural network that processes only the characters of that literal $\mathbf{I}_{::t}$, equation~\ref{eq:lit_embed} where $O[\mathbf{I}_{::t}]$ is the one-hot encoding of the characters. We use a gated recurrent unit (GRU)~\cite{gru} starting with $h_0 = \overrightarrow{\mathbf{0}}$ to process the characters in reverse order to emphasise the predicate and take the final hidden state $h_t$ to be the embedding of the literal $l \in \mathbb{R}^d$ where $d$ is the fixed dimension of the embedding. The context and query are embedded using the same network yielding a context tensor $\mathbf{C} \in \mathbb{R}^{R \times L \times d}$ where $R$ is the number of rules and $L$ number of literals in a rule; for the query we obtain vector $q \in \mathbb{R}^d$.

In order for the network to propagate the state unchanged, we append a \emph{null sentinel} $\phi = \overrightarrow{\mathbf{0}}$ which allows the network to ignore the current reasoning step by carrying over the state. We also append a \emph{blank rule} \texttt{()} that acts as a learnable parameter and is often attended when no other rule is suitable.

\subsection{Iteration}
The iterative step consists of attending to the rules, computing a new state using each rule and updating the old state. The network is iterated for $T$ steps, fixed in advance, with the initial state $s^0 \in \mathbb{R}^d$ set to the query vector $s^0 = q$.
\begin{align}
ca^t_i &= [s^t\ ;\ q\ ;\ r_i\ ;\ (s^t - r_i)^2\ ;\ s^t \odot r_i] \label{eq:att_feat}\\
\alpha^t_i &= \sigma ( W^{1 \times \frac{d}{2}} (U^{\frac{d}{2} \times d}ca^t_i + b^{\frac{d}{2}}) + b^1 ) \label{eq:att}
\end{align}
At any time step $t$, we compute a feature vector $ca^t_i \in \mathbb{R}^{5d}$ for every \emph{head} of a rule $r_i = \mathbf{C}_{i0}$ using the current state $s^t \in \mathbb{R}^d$, equation~\ref{eq:att_feat} where $[;]$ is the concatenation operator. We also experiment with embedding rules using another GRU over literals $h'_{ij} = GRU(\mathbf{C}_{ij}, h'_{i,j-1})$ and take the final state $r_i = h'_{iL}$ as the representation of the rule. To compute the final attention vector $\alpha^t_i$, we use a two layer feed-forward network, equation~\ref{eq:att} where $\sigma$ is the sigmoid function. We also experiment with the softmax formulation of the attention vector $\alpha^t_i$ after the two layer feed-forward network.
\begin{align}
u^t_{ij} &= GRU(\mathbf{C}_{ij}, u^t_{i(j-1)}) \label{eq:unifier} \\
s^{t+1} & = \sum^R_i \alpha^t_i u^t_{iL} \label{eq:update_state}
\end{align}
To \emph{apply} a rule, we use another recurrent neural network that processes every literal of every rule $\mathbf{C}_{ij}$. The initial hidden state $u^t_{i0} = s^t$ is set to the current state $s^t$, then for every rule a GRU is used to compute the new hidden state $u^t_{ij}$, equation~\ref{eq:unifier}. Finally, the new state $s^{t+1}$ becomes the weighted sum of the final hidden states $u^t_{iL}$, equation~\ref{eq:update_state}. We call the inner GRU unifier as it needs to learn unification between variables and constants as well as how each rule interacts with the current state.

\begin{figure*}[ht]
\centering
\includegraphics[width=\textwidth]{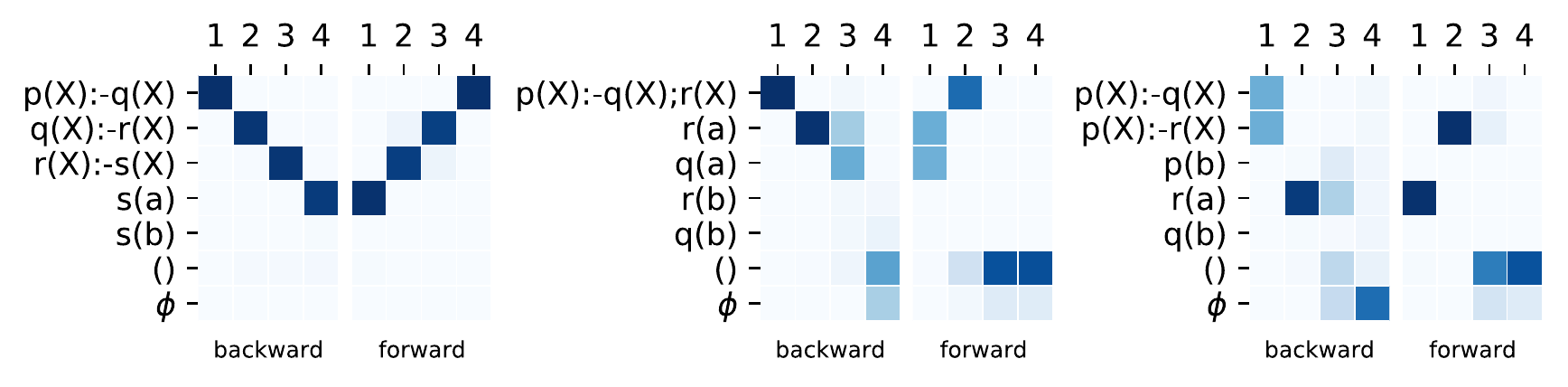}
\caption{Attention maps produced for query $p(a)$ for IMA with softmax attention performing \emph{backward} chaining in the left column and IMA with literal + rule embedding \emph{forward} chaining in the right column on tasks 5 to 7.}
\label{fig:attention}
\end{figure*}
\section{Experiments}\label{sec:experiments}
We carry out experiments on individual tasks with variations on our model. As a baseline, we use a LSTM~\cite{lstm} to process the query and then the context to predict the answer. We also compare our model against the Dynamic Memory Network (DMN)~\cite{dmn} and the Memory Attention Control (MAC)~\cite{mac} network which both incorporate iterative components achieving state-of-the-art results in visual question answering datasets. With DMN and MAC, the context is separated into rules and the entire rule is embedded using a GRU at the character level. Unlike DMN which uses another GRU to accumulate information over the rule embeddings and the current state, our variant IMA processes literal embeddings using a GRU to compute the new state as a weighted sum over each rule.

\textbf{Training} We use two training regimes: curriculum learning~\cite{curriculum} and multi-task. With curriculum learning the model is trained in an incremental fashion starting with tasks 1 and 2 with only 1 iteration. Then tasks accumulate with increasing number of iterations with tasks 3, 7, 9, 12 with 2 iterations and tasks 4, 6, 8, 11 using 3 iterations. We determine the minimum number iterations required for a task based on Prolog~\cite{prolog}. Finally all tasks are introduced with a maximum of 4 iterations. The multi-task approach trains on the entire dataset with 4 iterations fixed in advance. Models are trained via back-propagation using Adam~\cite{adam} for 120 epochs (10 per task) with a mini-batch size of 32 and a training dataset size of 20k logic programs per task. We ensure a mini-batch contains at least 1 sample from each training task to avoid any bias towards any task in a given mini-batch. Logic programs are shuffled after each epoch and rules within a context are also shuffled with every mini-batch. Since we have access to the data generating distribution, we do not use any regularisation in any of the models and have increased the training dataset size accordingly to avoid over-fitting~\cite{deeplearningbook}.

We generate 4 test sets of increasing difficulty: validation, easy, medium and hard which have up to 2, 4, 8 and 12 characters for predicates and constants as well as added number of irrelevant rules respectively. Each test set consists of 10k generated logic programs per task and results for the best single training run out of 3 for each model with state size $d=64$ on the easy set are shown in Table~\ref{tab:results}.

We observe that all iterative models perform better than the baseline except for task 8, transitivity which all models fail at. We speculate the reason is that the models have not seen existential variable binding for more than 2 character constants and fail to generalise in this particular case.
We note that the curriculum training regime has no benefit for any model most likely because we introduce new, unseen tasks with each iteration such that models with prior training have no advantage, ex. solving OR in 2 iterations does not improve the solution for AND in 3 iterations.
Embedding literals seems to provide an advantage over embedding rules since all IMA models outperform both DMN and MAC when we consider the mean accuracy over every test set. We postulate literal embeddings give a finer view and allow for better generalisation over increasing lengths as embedding rules with literals (lit+rule) also degrades the performance.
Although our variant IMA performs the best on all the test sets, all models quickly degrade in mean accuracy as the difficulty increases. We speculate that a finite sized state vector stores limited information about unseen unique sequences of increasing lengths and analyse this behaviour further in the next section. 

Figure~\ref{fig:attention} portrays the attention maps produced by IMA with softmax attention. Although by default models converge to \emph{backward} chaining, by (i) reversing the direction of the unifier GRU, equation~\ref{eq:unifier}, and (ii) skipping task 2, we can create a bias towards ground facts that have a matching constant. This approach encourages our model IMA with rule embeddings (lit+rule) to converge to \emph{forward} chaining~\cite{russell2016artificial}, albeit with more training time (thus results for forward chaining are not included in Table~\ref{tab:results}). This observation emphasises the fact that a fixed network architecture can be flexible enough to learn two different solutions for the same domain of problems given the right incentives.

\begin{figure}[h]
\centering
\includegraphics[width=0.9\columnwidth]{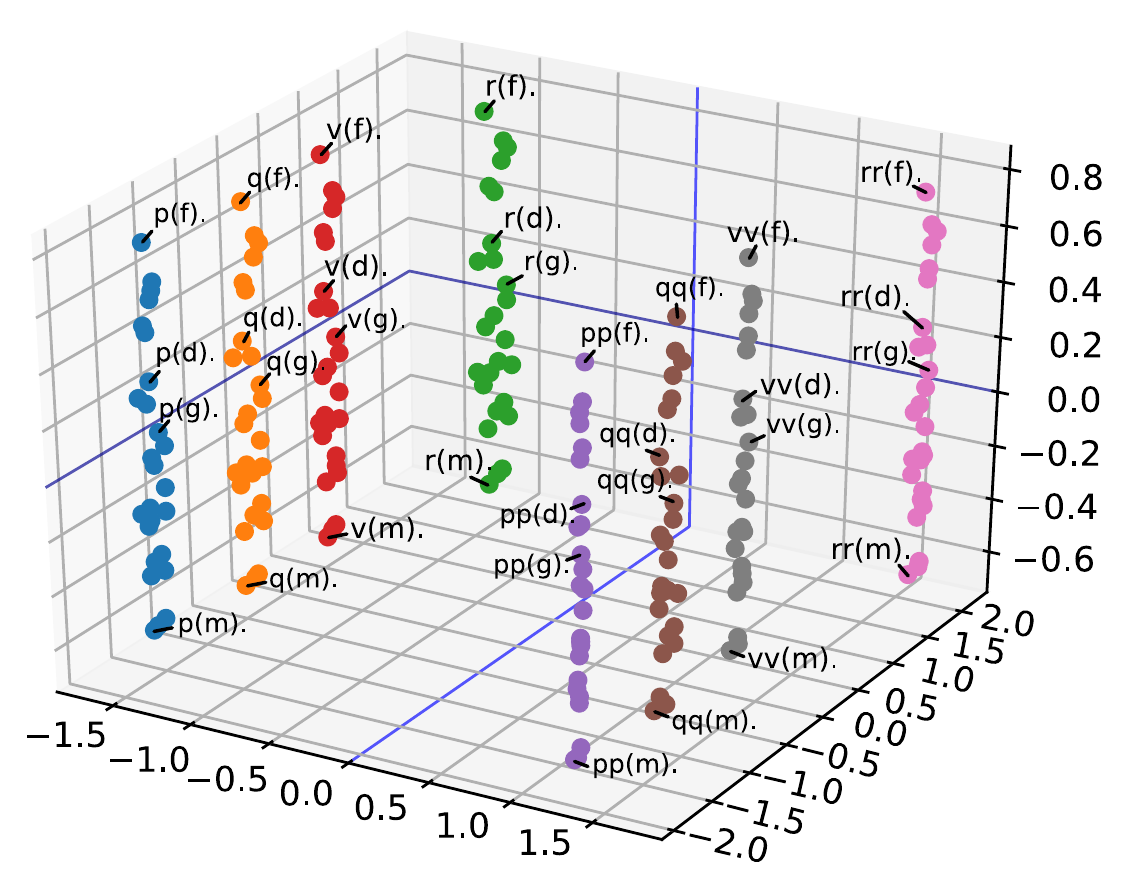}
\caption{First, second and last principal components of embeddings of single character literals form a lattice like structure with predicates clustered in vertical columns and constants on horizontal surfaces; from IMAsm.}
\label{fig:single_preds_imasm}
\end{figure}

\section{Analysis}\label{sec:analysis}
Since models are trained in an end-to-end fashion, the representations for logical constructs that help perform reasoning must also be learned. In this section, we scrutinise the learnt representations of the embedding GRU, equation~\ref{eq:lit_embed} for IMA and the rule embeddings of DMN.

An important assumption made in the dataset is that every predicate and constant combination is unique. We expect this setup to create a formation that would allow every literal to be differentiated. As such, we observe a lattice like structure when the embeddings of single and double character atoms are visualised in 3 dimensions using principle component analysis (PCA), Figure~\ref{fig:single_preds_imasm}. The first two components select the predicate and the final the constant to uniquely identify the literal with a clear distinction between single and double character predicates. This arrangement is in contrast with distributional representations that exploit similarities between entities such as in natural language embeddings~\cite{word2vec}. In our case, \texttt{p(a)} might be more \emph{similar} to \texttt{p(b)} than to \texttt{pp(a)} although all are deemed unique.
\begin{figure}[h]
\centering
\includegraphics[width=0.9\columnwidth]{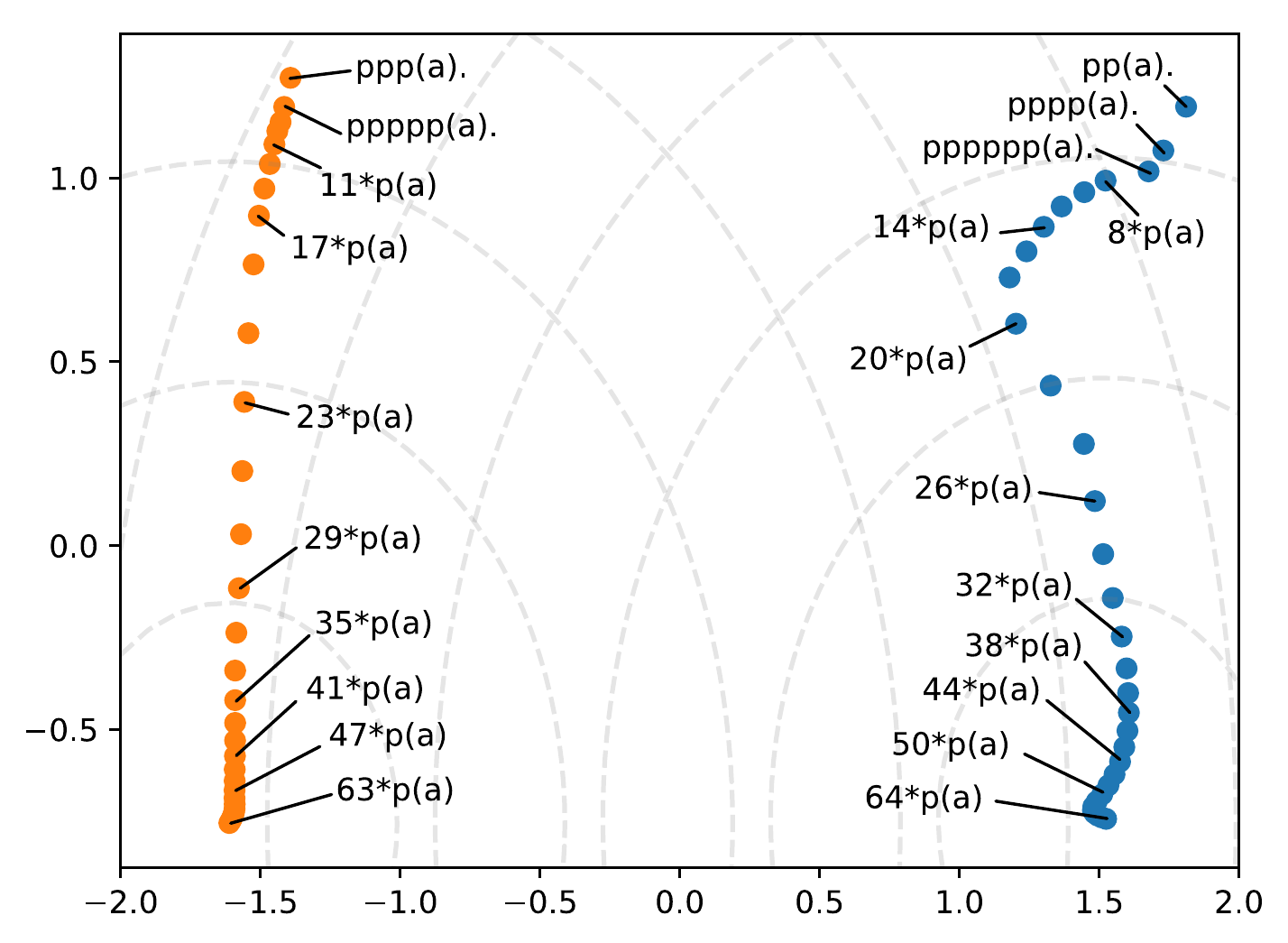}
\caption{Repeating character predicates saturate the embedding and converge to respective points, equidistant lines are plotted in grey; from IMAsm.}
\label{fig:pred_sat}
\end{figure}
\begin{figure*}[ht]
\centering
\begin{subfigure}[t]{0.47\textwidth}
\includegraphics[width=\textwidth]{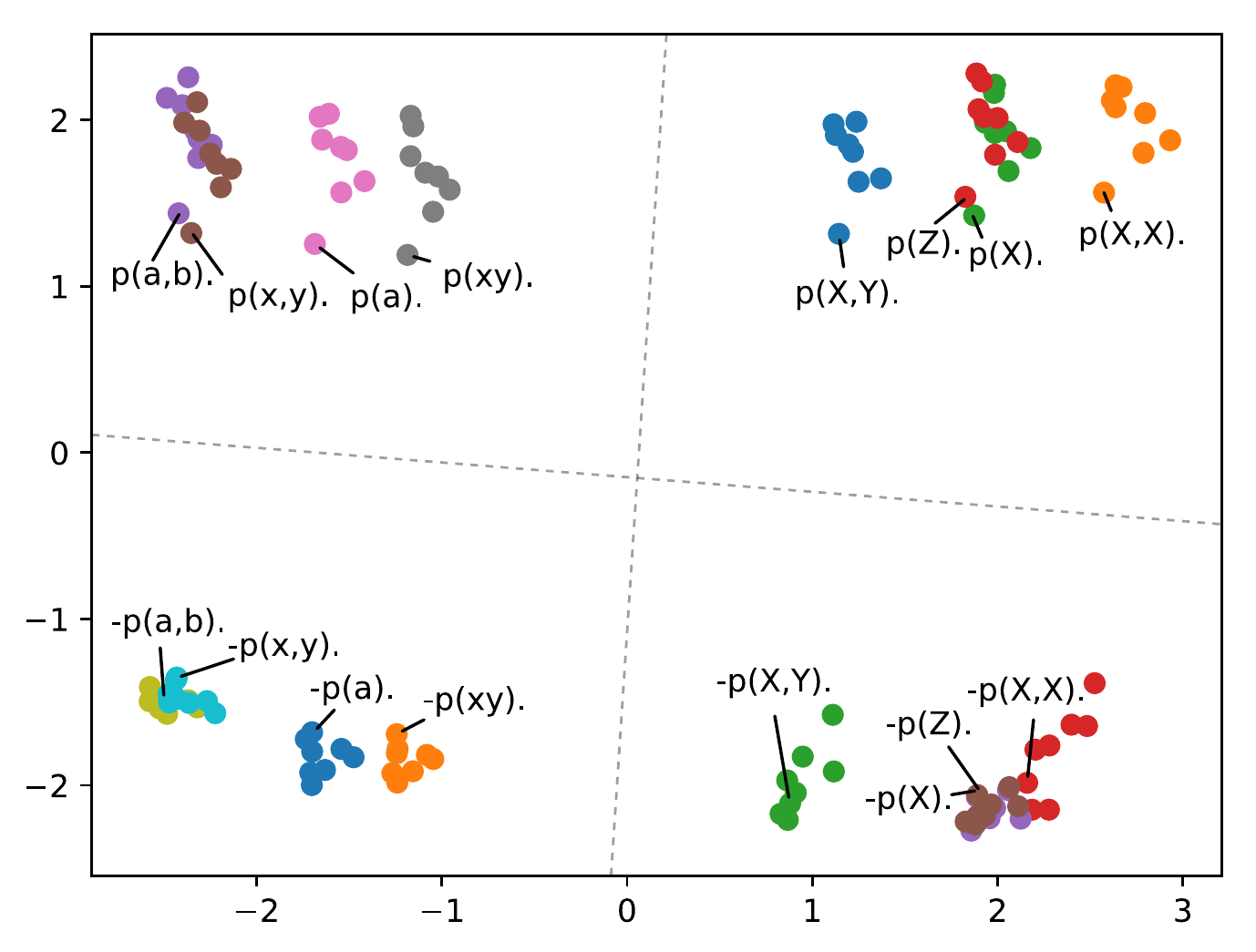}
\caption{Structurally different literals first cluster by whether they are negated or grounded then by arity (grey lines added as visual aids); from IMAsm.}
\label{fig:struct_preds_imasm}
\end{subfigure}
~
\begin{subfigure}[t]{0.47\textwidth}
\includegraphics[width=\textwidth]{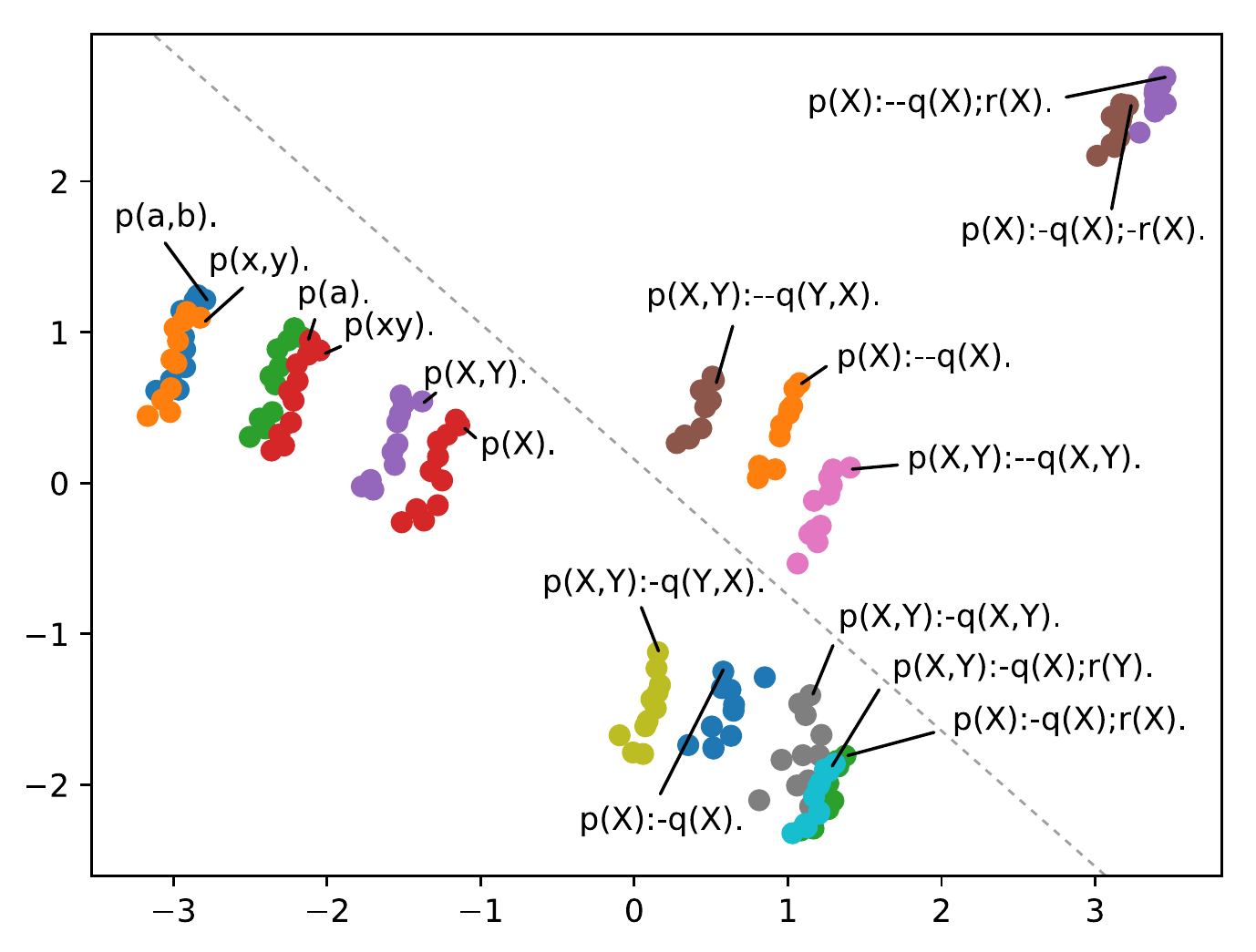}
\caption{Rule embeddings form clusters based on their structure with a distinction between negated and non-negated rules (grey line added as visual aid), from DMN.}
\label{fig:rules_dmn}
\end{subfigure}
\caption{Principle component analysis (PCA) of learnt representations.}
\label{fig:pcas}
\end{figure*}

As the embedding sizes are finite, we also expect a saturation in the embedding space as the length of the literals increase. We capture this phenomenon by repeating the character $p$ 64 times and observe a converging pattern shown in Figure~\ref{fig:pred_sat}. We take note how odd and even length predicates converge to their own respective point suggesting that the embeddings produced by the GRU, equation~\ref{eq:lit_embed}, learn parity.

If we take structurally different literals we observe a preference towards whether literals are negated or grounded then the arity of the atoms, Figure~\ref{fig:struct_preds_imasm}. We believe this clustering captures the literal semantics available in the dataset within 4 major clusters (grey lines). Furthermore, if we look at the rule embeddings learnt by DMN, we notice a similar clustering based on rule structure with again groundedness, negation, arity and number of body literals as learnt distinguishing features, Figure~\ref{fig:rules_dmn}. The multiple points within a cluster correspond to different predicates following a similar ordering, for example predicate \texttt{p} is often the upper most point.

\begin{figure}[h]
\centering
\includegraphics[width=0.9\columnwidth]{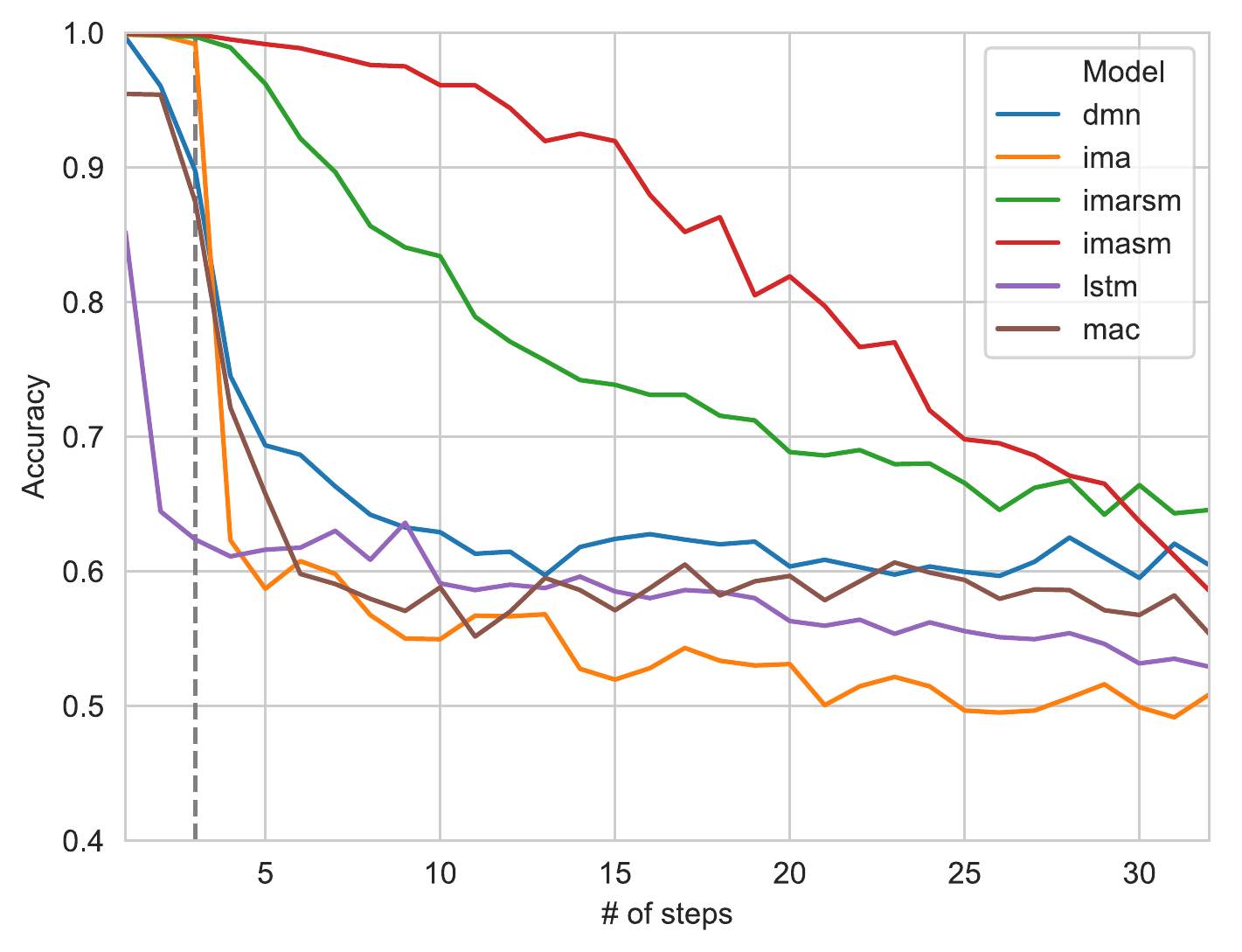}
\caption{When models are iterated beyond the training number of steps (grey line) to perform increasing steps of deduction, the accuracy degrades for all models.}
\label{fig:eval_nstep}
\end{figure}

To evaluate multi-hop reasoning, we generate logic programs that require increasing number of steps of deduction. While the training data contains up to 3 steps, we generate up to 32 steps and the networks are run for $n+1$ iterations. We obtain similar results to other recurrent neural network based systems~\cite{learnsimplealgo} and observe a steady decrease in accuracy beyond the training line (grey dashed line), shown in Figure~\ref{fig:eval_nstep} in which \texttt{imasm} and \texttt{imarsm} indicate IMA with softmax attention and rule embedding respectively. We speculate that with each iteration step the state representation degrades eventually losing information since models tend to produce noisy attention maps or state transformations. Our IMA model with softmax attention maintains the highest accuracy most likely due to the sparsity created by the softmax operation.

\begin{figure*}[ht]
\centering
\begin{subfigure}[t]{0.32\textwidth}
\includegraphics[width=\textwidth]{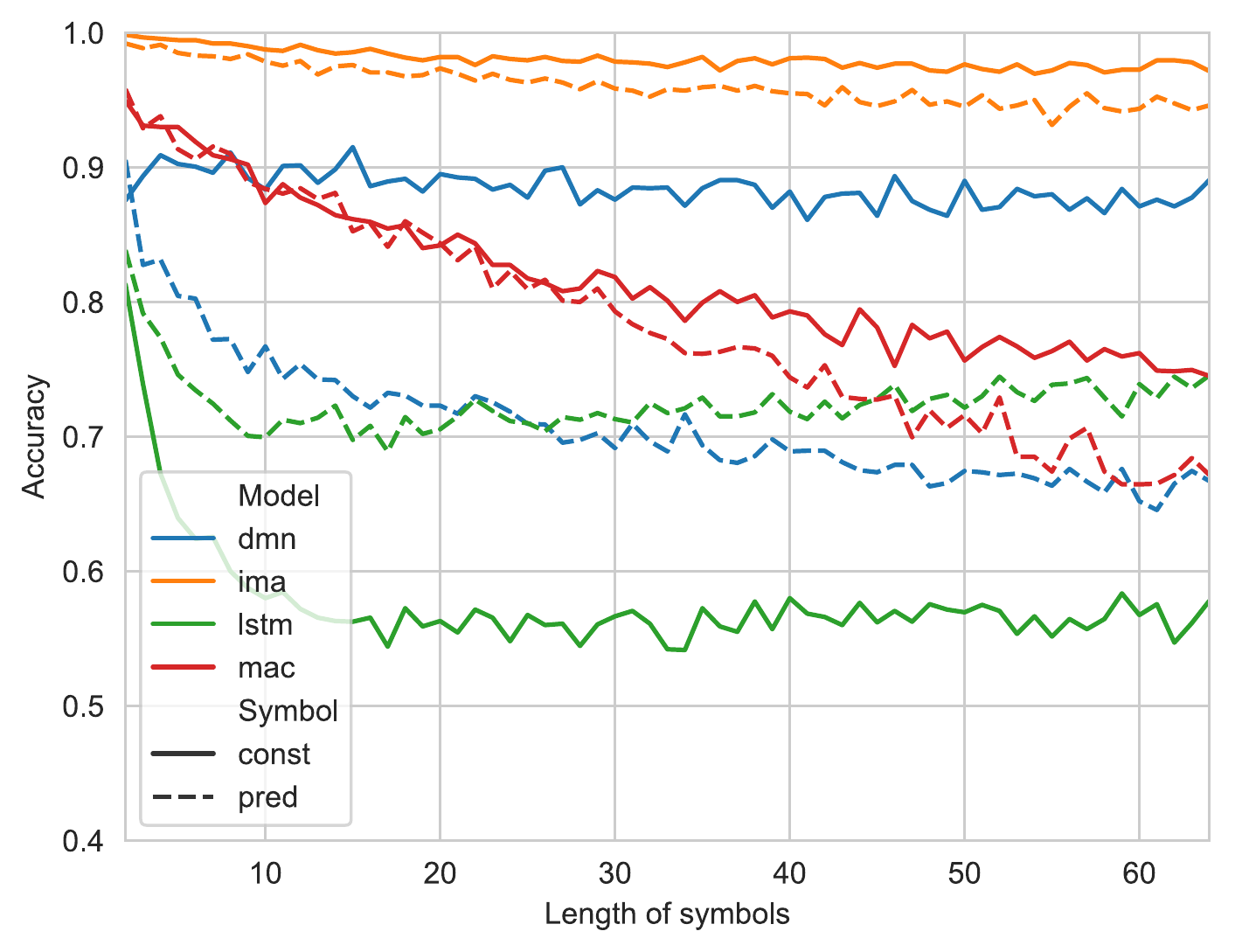}
\caption{The models can cope, in particular IMA with literal embeddings, when predicate and constant symbols of increasing length are randomly generated.}
\label{fig:eval_len}
\end{subfigure}
~
\begin{subfigure}[t]{0.32\textwidth}
\includegraphics[width=\textwidth]{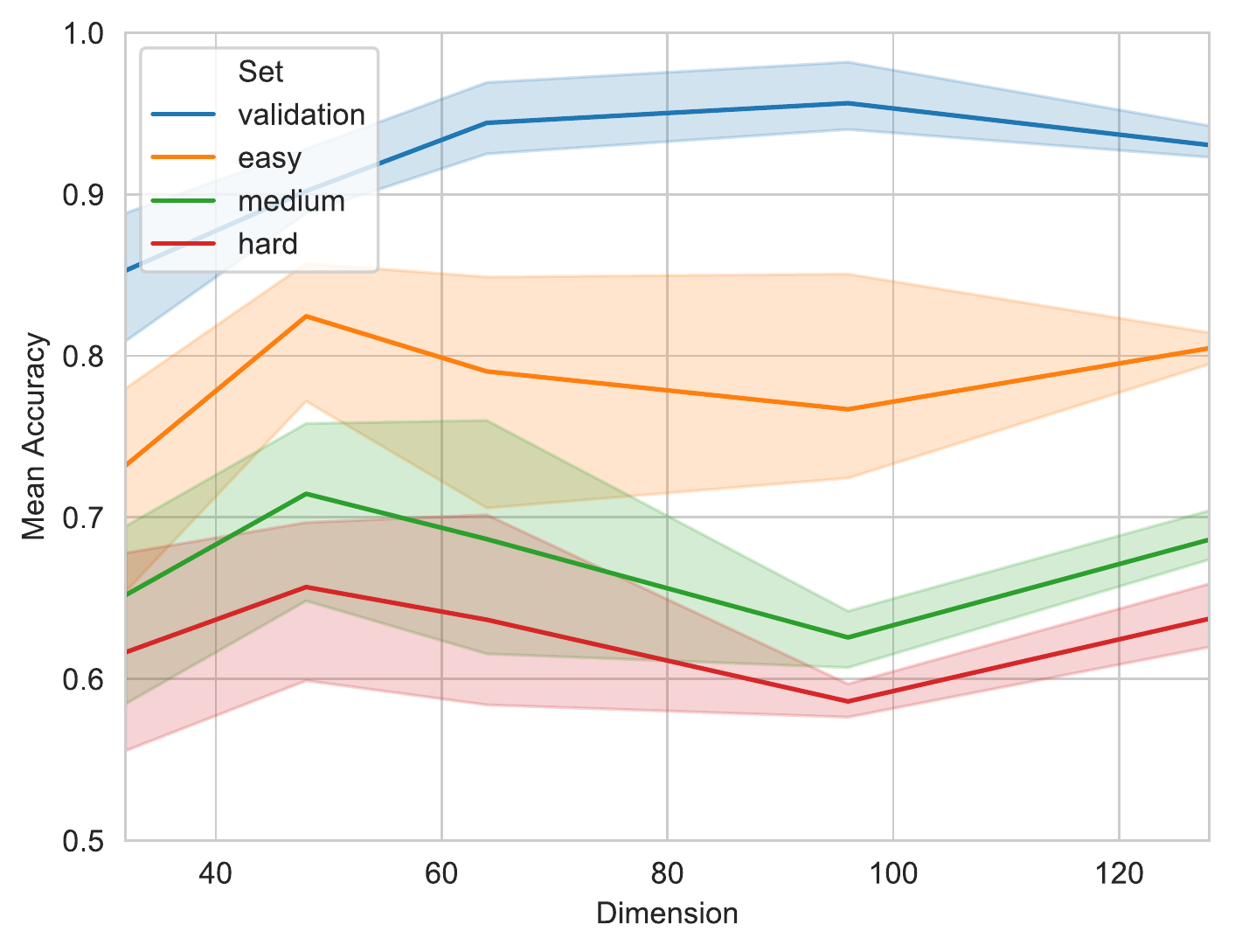}
\caption{Mean accuracy over all tasks against increasing embedding dimension $d$ shows no clear increase beyond $d=64$.}
\label{fig:eval_dim}
\end{subfigure}
~
\begin{subfigure}[t]{0.32\textwidth}
\includegraphics[width=\textwidth]{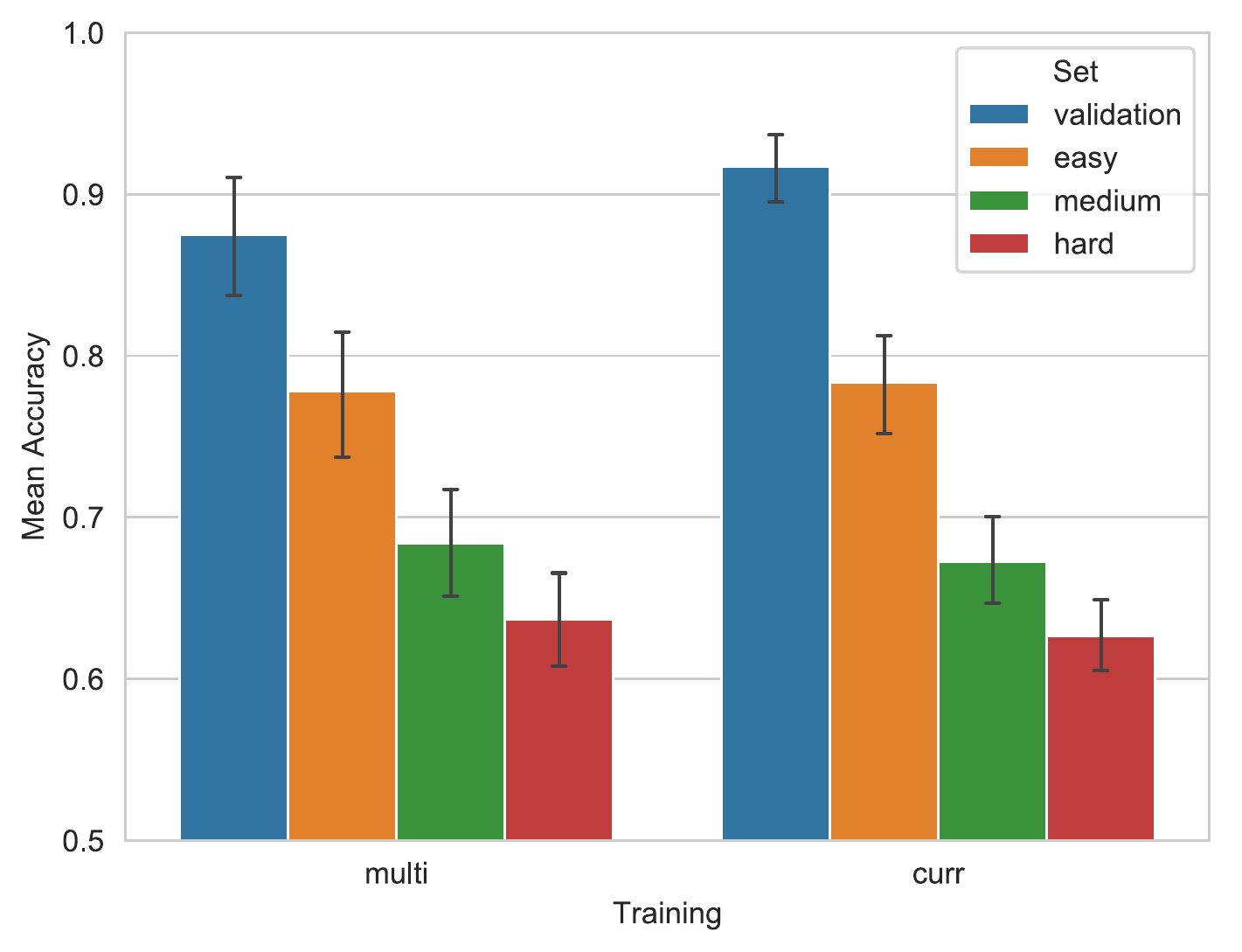}
\caption{Mean accuracy of training regimes applied to IMAsm against test sets across all dimensions show no advantage of curriculum training for generalisation.}
\label{fig:eval_training}
\end{subfigure}
\caption{Accuracy plots over increasing symbol lengths, state dimension and training regime.}
\label{fig:evals}
\end{figure*}

To evaluate generalisation to unseen symbols, we take task~3 and generate increasing character lengths of random predicate or constant symbols up to 64, well beyond the training dataset length of 2. Although we observed that literal embeddings can saturate, models can cope with longer randomly generated predicate or constant symbols, Figure~\ref{fig:eval_len}, since looking at only a few characters can determine uniqueness. This reflects on our intuition that looking at portions of sequences might be enough to determine equality rather than memorising them entirely, ex. looking at last few digits of phone numbers to check if they are same.

In order to understand how the embedding and state dimensions $d$ affect the models, we experimented with sizes 32, 48, 64, 96 and 128 running each curriculum training session 3 times for our IMA model. Figure~\ref{fig:eval_dim} shows that increasing the dimension size does not contribute to a clear increase in mean accuracy over all tasks and the drop in accuracy across easy, medium and hard test sets follow a similar pattern for every dimension. Despite the initial increase beyond $d=32$, we get the highest upper bound in accuracy with $d=64$, for which the individual results are in Table~\ref{tab:results}.

We designed the tasks to be of increasing difficulty building on previously seen problems such as unification before deduction. We expected models trained using an iterative scheme would generalise better as the network depth would increase gradually. However, when we average the mean accuracy for all dimensions, for all 3 runs of IMA with softmax attention across the test sets, we do not discern any advantage for curriculum training over multi-task training beyond the validation set. Figure~\ref{fig:eval_training} shows a similar decrease in performance across the test sets for both training regimes. We believe this result stems from introducing new tasks with each iteration and models not having any incentive to abstract subtasks until the next set of unseen tasks are incorporated into the training dataset.

\section{Discussion}
Given the simple, unit test like nature of the logic programs generated, the lack of robust results for iterative neural networks further encourages a combination of machine learning and symbolic systems rather than encompassing one with the other. In this section, we try to provide additional discussion and insight into why one might or might not learn symbolic reasoning from scratch using iterative neural networks.

Whilst the basic logic programs presented in this paper can \emph{all} be solved using existing symbolic systems such as Prolog~\cite{prolog}, we struggle to see a comparable performance from neural networks. Increasing the number of steps in task 3, as shown in Figure~\ref{fig:eval_nstep}, demonstrates the fragile nature of using continuous representations for rigid symbolic reasoning. The embedding space is inherently limited in capacity as it has fixed number of dimensions. Although a 64 dimensional vector can in theory encode a lot of information, in practice neural networks trained using back-propagation do not seem to have any guarantees on how efficiently they will learn representations in a vector space. We speculate that this creates an accumulative noise with each iteration which eventually degrades the performance. On the other hand, the learnt continuous representations scale to very long previously unseen symbols, Figure~\ref{fig:eval_len}, which is a desirable property of neuro-symbolic systems.

Attention mechanisms allow state-of-the-art memory based networks to address and operate an external memory~\cite{memnn}. In the case of nueral reasoning networks, the attention components provide the means to \emph{select} rules. An interesting outcome is that all neural reasoning models try to attend to multiple rules when applicable, for example when there are two matching rules as shown in Figure~\ref{fig:attention}. Unlike Prolog, this approach allows the neural networks to simultaneously explore multiple possible branches completing the reasoning task in fewer steps. However, we speculate such an attention mechanism will become the bottleneck when there are large numbers, possibly hundreds of matching rules such as in knowledge base completion tasks since it will try to aggregate all matching rules in a single step. An explicit strong supervision of the attention mechanism, similar to the original DMN~\cite{dmn} work or a hierarchical architecture might be needed to encourage a form of iterative backtracking.

Lack of abstraction between seemingly related tasks limits the performance on unseen logic programs. During curriculum training, we noticed models which solve tasks 1, 2 and 3 have no advantage on solving task 4 because the final state representation from task 3 specifically looks for a ground instance such as \texttt{p(a)} to complete the task. However, in task 4 the model needs to match another rule \texttt{p(X):-} and iterate once more. When presented with task 4, models most likely have to adjust the intermediate state representation to check for the second rule case; as a result Figure~\ref{fig:eval_training} can also be interpreted as the lack of abstraction over increasingly complex tasks in the dataset since curriculum learning provides no clear advantage. In other words, the neural models do not seem to learn a reasoning process that is general enough to be compared to existing symbolic systems; consequently we were unable to run them on existing logic program benchmarks or against Prolog.

\section{Related Work}\label{sec:related}
There have been many attempts to combine symbolic reasoning with machine learning techniques under the name neural-symbolic systems~\cite{neurosymbolic}. Lifted Relational Neural Networks~\cite{liftedneuralnetworks} ground clauses turning them into propositional programs when constructing networks as a set of weighted definite clauses. We do not pre-process programs to ground variables and Neural Reasoning Networks must learn unification. TensorLog~\cite{tensorlog} constructs ``factor graphs" from logic rules which in return create the network that run on one-hot encoded constants. Our approach does not factor, compile or use any implicit knowledge about first-order logical inference or rule applications. We also only one-hot encode characters, not entire predicates or constants, and only give labels 1 or 0 as targets to train end-to-end using the same neural network with same architecture and weights for every normal logic program. Logic Tensor Networks~\cite{logictensor} tackle knowledge completion ``on a simple but representative example" also grounding every term in the program prior to performing reasoning.

Following the above mentioned works, Neural Theorem Provers \cite{timntp} learn distributional representations of predicates and constants by symbolically constructing the relationship between embeddings using an existing symbolic inference engine. We design our tasks such that the neural networks attempt to learn not only representations at a character level but also the reasoning algorithm with no help or knowledge of existing reasoning engines and methods.

Learning similarities between constants can be used to perform link-prediction tasks~\cite{holographicembed} and knowledge base completion~\cite{neuraltensornetworks} but creates unwanted inferences when similar constants should indeed be unique~\cite{timlowdim}. Although we set every constant to be unique, we expect embeddings of similar constants to cluster during training if the data entails the same conclusions. Creating differentiable logical inference networks can also induce rules~\cite{evans2018learning}; however, at this stage we do not learn logical rules along side reasoning tasks and assume they are given to the model. Possible World Net~\cite{possibleworldnet} follows a unique approach to learning propositional logic entailment using semantic, worlds interpretation; however, they exploit the syntax of logical formulae by parsing them and constructing the neural network in a tree like manner in which nodes correspond to logical operators. The works mentioned so far are designed for either deductive databases, relation learning, link prediction, knowledge base completion or propositional programs; thus our task of learning reasoning over and embeddings of normal logic programs using a fixed RNN-based iterative network is inherently different making a direct empirical comparison unreasonable.

Neural Reasoning Networks can be seen as interpreters of logic programs as rules can act like instructions. This perspective reflects on systems such as Neural Program-Interpreters~\cite{neuralproginterpreters} and Neural Symbolic Machines~\cite{neuralsymbolicmachines}. These systems contain discrete operations that allow more complex actions to be performed overcoming the problem of a degrading state representation; however, they require a reinforcement learning setting to train. We believe a reinforcement learning approach applied on our dataset would learn a similar algorithm to that of Prolog~\cite{prolog}.

\section{Conclusion}\label{sec:conclusion}
We presented a new synthetic dataset and provided insights into how machine learning might encompass symbolic reasoning, defined as logic programs using RNN-based iterative neural networks. Fully differentiable models trained end-to-end have their inherent disadvantages: they seem to lose track when the number of iterations is increased and the embedding space is limited in capacity. However, such networks might still hold the key to incorporate symbolic prior knowledge into a continuous space by understanding how that embedding space is organised to store symbolic information.

Since such neural networks provide a differentiable but approximate reasoning engine over logic programs, in the future we hope to induce rules using continuous embeddings of logical rules by propagating gradients back to the context. However, initially if possible, a more robust neural reasoning network must be learned, one that is comparable in performance to existing logic-based expert systems. 

\bibliography{deeplogic}

\begin{thebibliography}{}

\bibitem[\protect\citeauthoryear{Apt and Bol}{1994}]{logicprogramming}
Apt, K.~R., and Bol, R.~N.
\newblock 1994.
\newblock Logic programming and negation: A survey.
\newblock {\em The Journal of Logic Programming} 19:9--71.

\bibitem[\protect\citeauthoryear{Apt and Van~Emden}{1982}]{bacwardchaining}
Apt, K.~R., and Van~Emden, M.~H.
\newblock 1982.
\newblock Contributions to the theory of logic programming.
\newblock {\em Journal of the ACM (JACM)} 29(3):841--862.

\bibitem[\protect\citeauthoryear{Battaglia \bgroup et al\mbox.\egroup
  }{2018}]{graphnetworks}
Battaglia, P.~W.; Hamrick, J.~B.; Bapst, V.; Sanchez-Gonzalez, A.; Zambaldi,
  V.; Malinowski, M.; Tacchetti, A.; Raposo, D.; Santoro, A.; Faulkner, R.;
  et~al.
\newblock 2018.
\newblock Relational inductive biases, deep learning, and graph networks.
\newblock {\em arXiv preprint arXiv:1806.01261}.

\bibitem[\protect\citeauthoryear{Bengio \bgroup et al\mbox.\egroup
  }{2009}]{curriculum}
Bengio, Y.; Louradour, J.; Collobert, R.; and Weston, J.
\newblock 2009.
\newblock Curriculum learning.
\newblock In {\em Proceedings of the 26th annual international conference on
  machine learning},  41--48.
\newblock ACM.

\bibitem[\protect\citeauthoryear{Besold \bgroup et al\mbox.\egroup
  }{2017}]{neurosymsurvey}
Besold, T.~R.; Garcez, A.~d.; Bader, S.; Bowman, H.; Domingos, P.; Hitzler, P.;
  K{\"u}hnberger, K.-U.; Lamb, L.~C.; Lowd, D.; Lima, P. M.~V.; et~al.
\newblock 2017.
\newblock Neural-symbolic learning and reasoning: A survey and interpretation.
\newblock {\em arXiv:1711.03902}.

\bibitem[\protect\citeauthoryear{Cho \bgroup et al\mbox.\egroup }{2014}]{gru}
Cho, K.; Van~Merri{\"e}nboer, B.; Bahdanau, D.; and Bengio, Y.
\newblock 2014.
\newblock On the properties of neural machine translation: Encoder-decoder
  approaches.
\newblock {\em arXiv:1409.1259}.

\bibitem[\protect\citeauthoryear{Clark}{1978}]{nbf}
Clark, K.~L.
\newblock 1978.
\newblock Negation as failure.
\newblock In {\em Logic and data bases}. Springer.
\newblock  293--322.

\bibitem[\protect\citeauthoryear{Clocksin and Mellish}{2003}]{prolog}
Clocksin, W.~F., and Mellish, C.~S.
\newblock 2003.
\newblock {\em Programming in PROLOG}.
\newblock Springer Science \& Business Media.

\bibitem[\protect\citeauthoryear{Cohen}{2016}]{tensorlog}
Cohen, W.~W.
\newblock 2016.
\newblock Tensorlog: A differentiable deductive database.
\newblock {\em arXiv:1605.06523}.

\bibitem[\protect\citeauthoryear{Ding}{1995}]{neuralprolog}
Ding, L.
\newblock 1995.
\newblock Neural prolog-the concepts, construction and mechanism.
\newblock In {\em Systems, Man and Cybernetics, 1995. Intelligent Systems for
  the 21st Century., IEEE International Conference on}, volume~4,  3603--3608.
\newblock IEEE.

\bibitem[\protect\citeauthoryear{Evans and
  Grefenstette}{2018}]{evans2018learning}
Evans, R., and Grefenstette, E.
\newblock 2018.
\newblock Learning explanatory rules from noisy data.
\newblock {\em Journal of Artificial Intelligence Research} 61:1--64.

\bibitem[\protect\citeauthoryear{Evans \bgroup et al\mbox.\egroup
  }{2018}]{possibleworldnet}
Evans, R.; Saxton, D.; Amos, D.; Kohli, P.; and Grefenstette, E.
\newblock 2018.
\newblock Can neural networks understand logical entailment?
\newblock {\em arXiv:1802.08535}.

\bibitem[\protect\citeauthoryear{Garcez, Broda, and
  Gabbay}{2012}]{neurosymbolic}
Garcez, A. S.~d.; Broda, K.~B.; and Gabbay, D.~M.
\newblock 2012.
\newblock {\em Neural-symbolic learning systems: foundations and applications}.
\newblock Springer Science \& Business Media.

\bibitem[\protect\citeauthoryear{Goodfellow, Bengio, and
  Courville}{2017}]{deeplearningbook}
Goodfellow, I.; Bengio, Y.; and Courville, A.
\newblock 2017.
\newblock {\em Deep Learning}.
\newblock The MIT Press.

\bibitem[\protect\citeauthoryear{Graves \bgroup et al\mbox.\egroup
  }{2016}]{dnc}
Graves, A.; Wayne, G.; Reynolds, M.; Harley, T.; Danihelka, I.;
  Grabska-Barwi{\'n}ska, A.; Colmenarejo, S.~G.; Grefenstette, E.; Ramalho, T.;
  Agapiou, J.; et~al.
\newblock 2016.
\newblock Hybrid computing using a neural network with dynamic external memory.
\newblock {\em Nature} 538(7626):471.

\bibitem[\protect\citeauthoryear{Graves, Wayne, and
  Danihelka}{2014}]{neuralturing}
Graves, A.; Wayne, G.; and Danihelka, I.
\newblock 2014.
\newblock Neural turing machines.
\newblock {\em arXiv:1410.5401}.

\bibitem[\protect\citeauthoryear{Hochreiter and Schmidhuber}{1997}]{lstm}
Hochreiter, S., and Schmidhuber, J.
\newblock 1997.
\newblock Long short-term memory.
\newblock {\em Neural Computation} 9(8):1735--1780.

\bibitem[\protect\citeauthoryear{Hudson and Manning}{2018}]{mac}
Hudson, D.~A., and Manning, C.~D.
\newblock 2018.
\newblock Compositional attention networks for machine reasoning.
\newblock {\em arXiv:1803.03067}.

\bibitem[\protect\citeauthoryear{Kingma and Ba}{2014}]{adam}
Kingma, D.~P., and Ba, J.
\newblock 2014.
\newblock Adam: A method for stochastic optimization.
\newblock {\em arXiv:1412.6980}.

\bibitem[\protect\citeauthoryear{Komendantskaya}{2011}]{neuralunification}
Komendantskaya, E.
\newblock 2011.
\newblock Unification neural networks: unification by error-correction
  learning.
\newblock {\em Logic Journal of the IGPL} 19(6):821--847.

\bibitem[\protect\citeauthoryear{Kumar \bgroup et al\mbox.\egroup }{2016}]{dmn}
Kumar, A.; Irsoy, O.; Ondruska, P.; Iyyer, M.; Bradbury, J.; Gulrajani, I.;
  Zhong, V.; Paulus, R.; and Socher, R.
\newblock 2016.
\newblock Ask me anything: Dynamic memory networks for natural language
  processing.
\newblock In {\em ICML},  1378--1387.

\bibitem[\protect\citeauthoryear{Liang \bgroup et al\mbox.\egroup
  }{2016}]{neuralsymbolicmachines}
Liang, C.; Berant, J.; Le, Q.; Forbus, K.~D.; and Lao, N.
\newblock 2016.
\newblock Neural symbolic machines: Learning semantic parsers on freebase with
  weak supervision.
\newblock {\em arXiv:1611.00020}.

\bibitem[\protect\citeauthoryear{Mikolov \bgroup et al\mbox.\egroup
  }{2013}]{word2vec}
Mikolov, T.; Sutskever, I.; Chen, K.; Corrado, G.~S.; and Dean, J.
\newblock 2013.
\newblock Distributed representations of words and phrases and their
  compositionality.
\newblock In {\em NIPS},  3111--3119.

\bibitem[\protect\citeauthoryear{Muggleton and De~Raedt}{1994}]{ilp}
Muggleton, S., and De~Raedt, L.
\newblock 1994.
\newblock Inductive logic programming: Theory and methods.
\newblock {\em The Journal of Logic Programming} 19:629--679.

\bibitem[\protect\citeauthoryear{Nickel \bgroup et al\mbox.\egroup
  }{2016}]{holographicembed}
Nickel, M.; Rosasco, L.; Poggio, T.~A.; et~al.
\newblock 2016.
\newblock Holographic embeddings of knowledge graphs.
\newblock In {\em AAAI},  1955--1961.

\bibitem[\protect\citeauthoryear{Pennington, Socher, and Manning}{2014}]{glove}
Pennington, J.; Socher, R.; and Manning, C.
\newblock 2014.
\newblock Glove: Global vectors for word representation.
\newblock In {\em EMNLP},  1532--1543.

\bibitem[\protect\citeauthoryear{Reed and
  De~Freitas}{2015}]{neuralproginterpreters}
Reed, S., and De~Freitas, N.
\newblock 2015.
\newblock Neural programmer-interpreters.
\newblock {\em arXiv:1511.06279}.

\bibitem[\protect\citeauthoryear{Rockt{\"a}schel and Riedel}{2017}]{timntp}
Rockt{\"a}schel, T., and Riedel, S.
\newblock 2017.
\newblock End-to-end differentiable proving.
\newblock In {\em NIPS},  3791--3803.

\bibitem[\protect\citeauthoryear{Rockt{\"a}schel \bgroup et al\mbox.\egroup
  }{2014}]{timlowdim}
Rockt{\"a}schel, T.; Bo{\v{s}}njak, M.; Singh, S.; and Riedel, S.
\newblock 2014.
\newblock Low-dimensional embeddings of logic.
\newblock In {\em Proceedings of the ACL 2014 Workshop on Semantic Parsing},
  45--49.

\bibitem[\protect\citeauthoryear{Russell and
  Norvig}{2016}]{russell2016artificial}
Russell, S., and Norvig, P.
\newblock 2016.
\newblock {\em Artificial Intelligence: A Modern Approach (3rd Edition)}.
\newblock Pearson.

\bibitem[\protect\citeauthoryear{Santoro \bgroup et al\mbox.\egroup
  }{2017}]{relationnetworks}
Santoro, A.; Raposo, D.; Barrett, D.~G.; Malinowski, M.; Pascanu, R.;
  Battaglia, P.; and Lillicrap, T.
\newblock 2017.
\newblock A simple neural network module for relational reasoning.
\newblock In {\em NIPS},  4974--4983.

\bibitem[\protect\citeauthoryear{Serafini and Garcez}{2016}]{logictensor}
Serafini, L., and Garcez, A.~d.
\newblock 2016.
\newblock Logic tensor networks: Deep learning and logical reasoning from data
  and knowledge.
\newblock {\em arXiv:1606.04422}.

\bibitem[\protect\citeauthoryear{Socher \bgroup et al\mbox.\egroup
  }{2013}]{neuraltensornetworks}
Socher, R.; Chen, D.; Manning, C.~D.; and Ng, A.
\newblock 2013.
\newblock Reasoning with neural tensor networks for knowledge base completion.
\newblock In {\em NIPS},  926--934.

\bibitem[\protect\citeauthoryear{Sourek \bgroup et al\mbox.\egroup
  }{2015}]{liftedneuralnetworks}
Sourek, G.; Aschenbrenner, V.; Zelezny, F.; and Kuzelka, O.
\newblock 2015.
\newblock Lifted relational neural networks.
\newblock {\em arXiv:1508.05128}.

\bibitem[\protect\citeauthoryear{Sukhbaatar \bgroup et al\mbox.\egroup
  }{2015}]{memn2n}
Sukhbaatar, S.; Weston, J.; Fergus, R.; et~al.
\newblock 2015.
\newblock End-to-end memory networks.
\newblock In {\em NIPS},  2440--2448.

\bibitem[\protect\citeauthoryear{Tai, Socher, and Manning}{2015}]{treelstm}
Tai, K.~S.; Socher, R.; and Manning, C.~D.
\newblock 2015.
\newblock Improved semantic representations from tree-structured long
  short-term memory networks.
\newblock {\em arXiv preprint arXiv:1503.00075}.

\bibitem[\protect\citeauthoryear{Weston \bgroup et al\mbox.\egroup
  }{2015}]{babi}
Weston, J.; Bordes, A.; Chopra, S.; Rush, A.~M.; van Merri{\"e}nboer, B.;
  Joulin, A.; and Mikolov, T.
\newblock 2015.
\newblock Towards ai-complete question answering: A set of prerequisite toy
  tasks.
\newblock {\em arXiv:1502.05698}.

\bibitem[\protect\citeauthoryear{Weston, Chopra, and Bordes}{2015}]{memnn}
Weston, J.; Chopra, S.; and Bordes, A.
\newblock 2015.
\newblock Memory networks.
\newblock {\em ICLR}.

\bibitem[\protect\citeauthoryear{Xiong, Meity, and Socher}{2016}]{dmnvisual}
Xiong, C.; Meity, S.; and Socher, R.
\newblock 2016.
\newblock Dynamic memory networks for visual and textual question answering.
\newblock In {\em ICML},  2397--2406.

\bibitem[\protect\citeauthoryear{Zaremba \bgroup et al\mbox.\egroup
  }{2016}]{learnsimplealgo}
Zaremba, W.; Mikolov, T.; Joulin, A.; and Fergus, R.
\newblock 2016.
\newblock Learning simple algorithms from examples.
\newblock In {\em ICML},  421--429.

\end{thebibliography}
\bibliographystyle{aaai}
\end{document}